\documentclass[10pt,twocolumn,letterpaper]{article}
\usepackage[accsupp]{axessibility} % Improves PDF readability for those with disabilities.

\usepackage{cvpr}              % To produce the CAMERA-READY version
\definecolor{cvprblue}{rgb}{0.21,0.49,0.74}
\definecolor{mygray}{gray}{.9}
\usepackage[pagebackref,breaklinks,colorlinks,allcolors=cvprblue]{hyperref}

\def\confName{CVPR}
\def\confYear{2025}

\title{Complementary Advantages: 
Exploiting Cross-Field Frequency Correlation for NIR-Assisted Image Denoising}
\usepackage{multirow}
\usepackage{subcaption}
\usepackage{colortbl}
\usepackage{float}
\usepackage{graphicx} 
\def\authorBlock{
    Yuchen Wang$^{1} $ \qquad Hongyuan Wang$^{1}$ \qquad Lizhi Wang$^{2, 3}$\thanks{Corresponding Author: Lizhi Wang (wanglizhi@bnu.edu.cn)} \qquad\\
    Xin Wang$^{1}$ \qquad Lin Zhu$^{1}$ \qquad Wanxuan Lu$^{4}$ \qquad Hua Huang$^{2, 3}$\\
    $^{1}$ School of Computer Science and Technology, Beijing Institute of Technology \\
    $^{2}$ School of Artificial Intelligence, Beijing Normal University \\
    $^{3}$ Engineering Research Center of Intelligent Technology and Educational Application, Ministry of Education \\
    $^{4}$ Aerospace Information Research Institute, Chinese Academy of Sciences\\}
\author{\authorBlock}

\begin{document}
\maketitle
\begin{abstract}
Existing single-image denoising algorithms often struggle to restore details when dealing with complex noisy images. The introduction of near-infrared (NIR) images offers new possibilities for RGB image denoising. However, due to the inconsistency between NIR and RGB images, the existing works still struggle to balance the contributions of two fields in the process of image fusion. In response to this, in this paper, we develop a cross-field Frequency Correlation Exploiting Network (FCENet) for NIR-assisted image denoising. We first propose the frequency correlation prior based on an in-depth statistical frequency analysis of NIR-RGB image pairs. The prior reveals the complementary correlation of NIR and RGB images in the frequency domain. Leveraging frequency correlation prior, we then establish a frequency learning framework composed of Frequency Dynamic Selection Mechanism (FDSM) and Frequency Exhaustive Fusion Mechanism (FEFM). FDSM dynamically selects complementary information from NIR and RGB images in the frequency domain, and FEFM strengthens the control of common and differential features during the fusion process of NIR and RGB features. Extensive experiments on simulated and real data validate that the proposed method outperforms other state-of-the-art methods. The code will be released at \href{https://github.com/yuchenwang815/FCENet}{https://github.com/yuchenwang815/FCENet}.
\end{abstract}    
\section{Introduction}
% \label{sec:intro}
Due to the inherent physical limitations of digital imaging devices, the captured images are often affected by various types of noise, which have a considerable impact on applications such as 24-hour surveillance, autonomous driving, and smartphone photography. Although the development of deep learning techniques~\cite{vaswani2017attention,liang2021swinir} has brought forth a variety of image denoising methods\begin{figure}[t!]
    \begin{center}
        \includegraphics[width=\linewidth]{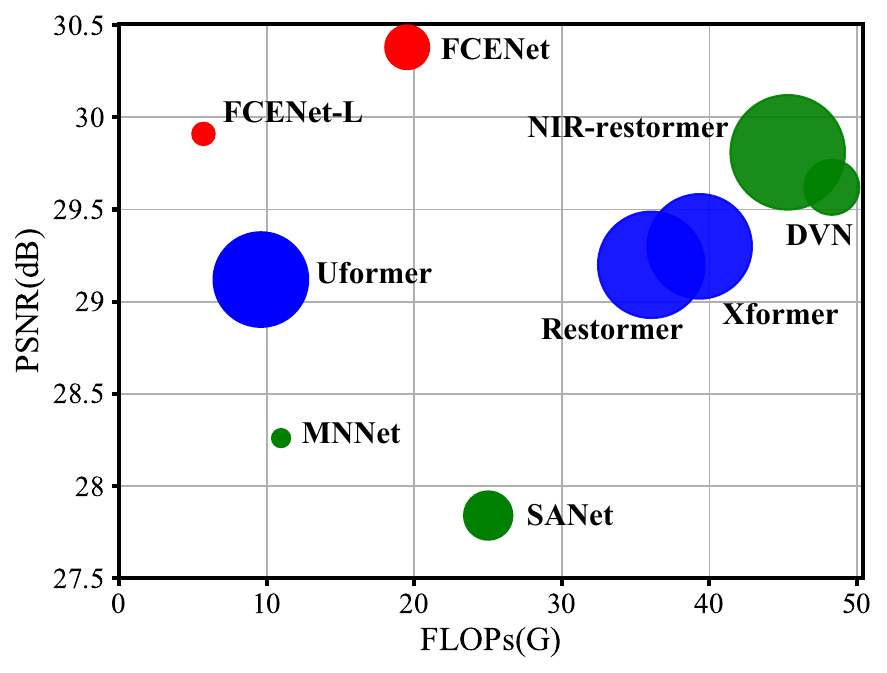}
    \end{center}
    \vspace{-6mm}
    \caption{Comparisons of PSNR, FLOPs, and Parameters on the DVD dataset~\cite{jin2022darkvisionnet} are presented. Green circles represent the single-image denoising algorithms, and blue circles represent the NIR-assisted image denoising algorithms. The circle radius represents the number of parameters. The proposed method (red circles) achieves the superior performance while maintaining efficiency.}
    \label{fig:1}
\vspace{-4mm}
\end{figure}~\cite{abdelhamed2020ntire,zamir2021multi,feng2022learnability,zhang2018ffdnet,batson2019noise2self,mansour2023zero}, existing algorithms still find it difficult to deal with complex noisy images, especially in low-light environment. Restoring clean detailed information from noisy images that lack substantial information is highly ill-posed. Therefore, it is necessary to introduce supplementary information from new fields. Fortunately, NIR images with high Signal to Noise Ratio (SNR) can be captured at a low cost and are suitable for everyday photography. Thus, NIR-assisted image denoising~\cite{yan2013cross,lv2020integrated,9999306} has become a promising solution for image denoising.

\begin{figure}[t!]
    \begin{center}
        \includegraphics[width=\linewidth]{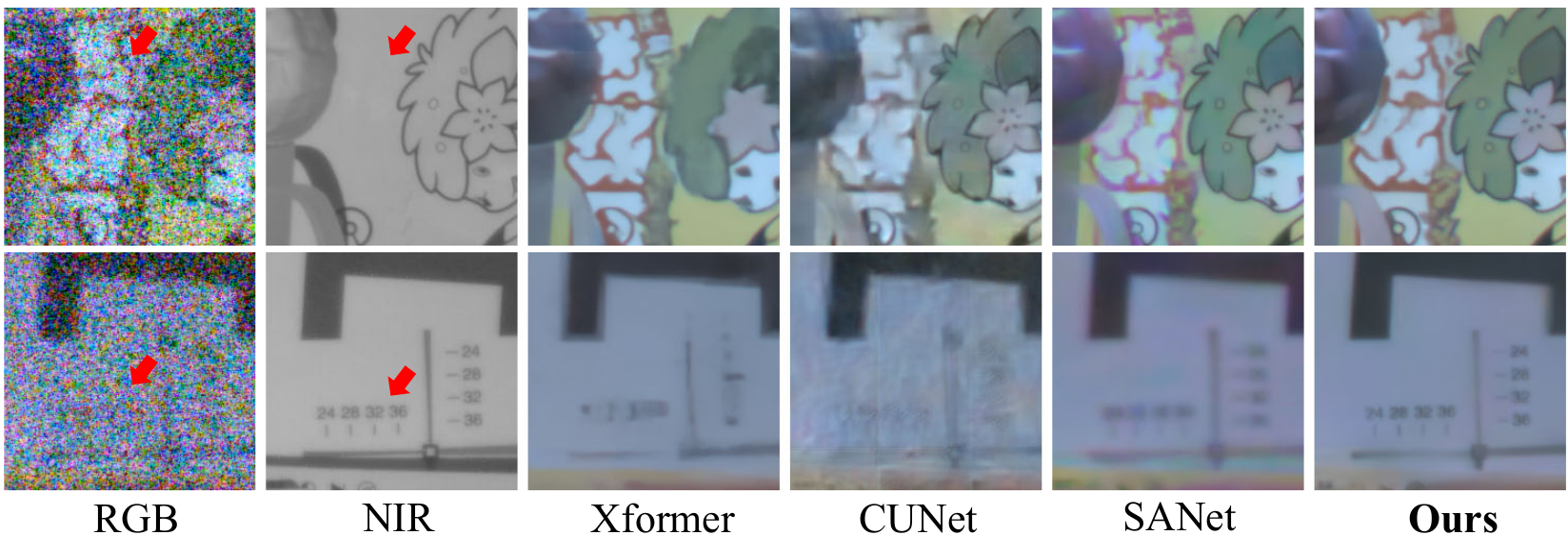}
    \end{center}
    \vspace{-6mm}
    \caption{ Visual comparisons on the challenging noisy RGB-NIR image pairs. Pay attention to the differences in structural and color information between two fields. Our method produces a better denoising result with clear details and fewer artifacts.}
    \label{fig:2}
\vspace{-5mm}
\end{figure}

However, it is challenging to effectively leverage information from different
fields to achieve noise reduction. Due to the influence of the inherent reflective spectra of objects, there are inconsistencies between NIR and RGB images. As shown in Fig.~\ref{fig:2}, firstly, there is a significant difference in color and brightness information between two fields. Secondly, some structural texture of RGB image disappears in NIR image, and additional artifacts may appear. The color and structural inconsistencies between NIR and RGB images pose a significant challenge for the task of NIR-assisted image denoising.

In existing researches, some methods attempt to directly integrate inconsistent features from two fields~\cite{xu2022model,xu2024nir,sheng2022frequency,sheng2023structure}. However, these methods indiscriminately fuse cross-field information, relying on the fitting capabilities of neural networks to learn the fusion of NIR and RGB images under spatial inconsistency, which often leads to suboptimal denoising results and lower computational efficiency. Some methods reinforce common features from NIR and RGB images and diminish the weight of inconsistent features during the fusion~\cite{yan2013cross,kim2021deformable,deng2020deep,jin2022darkvisionnet}. These image fusion schemes typically only enhance the common information in NIR and RGB during fusion, but some differential information such as the sharp textures in NIR and the color information in RGB is discarded.

In this paper, we develop a cross-field Frequency Correlation Exploiting Network (FCENet) for NIR-assisted image denoising. We conduct a series of analyses on the characteristics of NIR and RGB images in the frequency domain and obtain a key observation: in the same scene, the similarity between noisy RGB images and clean RGB (noise-free RGB) images decreases from low to high frequency, whereas the similarity between NIR images and clean RGB images increases from low to high frequency. This significant cross-field frequency correlation prior redefines cross-field fusion as a process of frequency domain adaptive selection, motivating us to subsequently establish detailed frequency domain learning to precisely control the restoration of texture details and colors. 

Specifically, the frequency domain fusion network consists of two mechanisms: Frequency Dynamic Selection Mechanism (FDSM) and Frequency Exhaustive Fusion Mechanism (FEFM). In order to comprehensively extract useful features from inconsistent images, the FDSM is formulated based on dynamic filters under the guidance of frequency domain prior. The FDSM efficiently screens complementary features from NIR and RGB in the frequency domain, providing high quality material for subsequent image fusion. Then, the FEFM is formulated to ensure thorough frequency domain fusion. The FEFM integrates local and long-range correlations across fields to fully extract the common feature from cross-field information, and employs a differential cross-attention approach to complement some key differential high frequency features in NIR features. 

The proposed FCENet can utilize complementary information from cross-field images. Extensive experiments on simulated and real data validate that our method outperforms other state-of-the-art methods in terms of image quality and computational efficiency. The main contributions of this work are summarized as follows:
\begin{itemize}
    \item We establish the frequency correlation prior between NIR and RGB images through a series of frequency domain analyses of two fields, which prompts the integration of complementary features in the frequency domain.
    \item We propose an efficient frequency dynamic selection mechanism based on the frequency correlation prior to identify which useful low-frequency and high frequency information should be retained in NIR and RGB features.
    \item We develop an exhaustive frequency domain fusion mechanism that models the local similarity and long-range correlation in the frequency domain of NIR and RGB images, and supplements some critical high frequency differential features.
\end{itemize}

\section{Related Work}
% \label{sec:formatting}
\subsection{Single Image Denoising}
Image denoising~\cite{portilla2003image,elad2006image} is a classic task in low-level vision. In recent years, deep learning has greatly improved the accuracy and efficiency of denoising~\cite{anwar2019real,cheng2021nbnet,tian2020deep,li2024stimulating,feng2023learnability,li2024positive2negative_cvpr}. DnCNN~\cite{7839189} uses a simple network with batch normalization and residual learning, which outperforms traditional denoising methods. MPRNet~\cite{zamir2021multi} designs a multi-stage restoration framework, and Restormer~\cite{zamir2022restormer} proposes an efficient transformer framework for high-resolution images. Xformer~\cite{zhang2023xformer} proposes a transformer framework that combines spatial self-attention and channel self-attention. However, despite their strong denoising capabilities, single-image denoising still inevitably loses a lot of spatial information, leading to overly smooth edge textures in the denoised images.
\subsection{NIR-Assisted Image Restoration}
To achieve better denoising effects, researchers have attempted to use NIR images to assist in RGB image denoising~\cite{yan2013cross,lv2020integrated,cheng2023mutually,wang2019stereoscopic}. Compared with single-image recovery, NIR images can help restore the details of degraded images~\cite{xu2024nir,9999306}. In early research, Yan et al.~\cite{yan2013cross} proposes a scale map from the perspective of gradient inconsistency to find universal usable edges and smooth transitions in NIR images. With the proliferation of deep learning, CUNet~\cite{deng2020deep} uses a convolutional sparse coding model to extract common features from cross-field images. FGDNet~\cite{sheng2022frequency} and SANet~\cite{sheng2023structure} fuse the features of two fields in the frequency domain. MNNet~\cite{xu2022model} proposes an observation model that takes into account the modality gap between the target and guiding images. DVN~\cite{jin2022darkvisionnet} integrates structural inconsistency priors into deep features, specifically strengthening the common deep features in NIR and RGB images. NAID~\cite{xu2024nir} proposes Selective Fusion Module (SFM), which can be plugged into advanced image restoration networks. However, under extreme noisy conditions, existing NIR-assisted denoising methods fail to balance color accuracy, detail retention, and artifact suppression.
\subsection{Frequency Domain Image Restoration}
Frequency analysis in digital signal processing emphasizes the frequency properties of signals~\cite{tancik2020fourier,xu2019frequency}. Recently, frequency domain analysis has shown great potential in improving the performance of various learning-based image restoration frameworks~\cite{jiang2021focal,zou2021sdwnet,wei2021unsupervised}. In~\cite{kong2023efficient}, An efficient estimation of scaled dot-product attention based on a frequency domain self-attention solver (FSAS) is proposed. HDNet~\cite{hu2022hdnet} proposes a high-resolution dual-domain learning network. SFNet~\cite{cui2023selective} uses filters to dynamically decouple feature mappings into different frequency components and adaptively extracts useful features. In this paper, we emphasize the importance of frequency domain information for the research on fusion under inconsistency and design a comprehensive frequency domain fusion scheme for this task to ensure the utilization of complementary information from NIR and RGB images.

\section{Method}
In this section, \begin{figure}[t]
  \centering
  \begin{subfigure}[b]{0.23\textwidth}
    \centering
    \includegraphics[width=\linewidth]{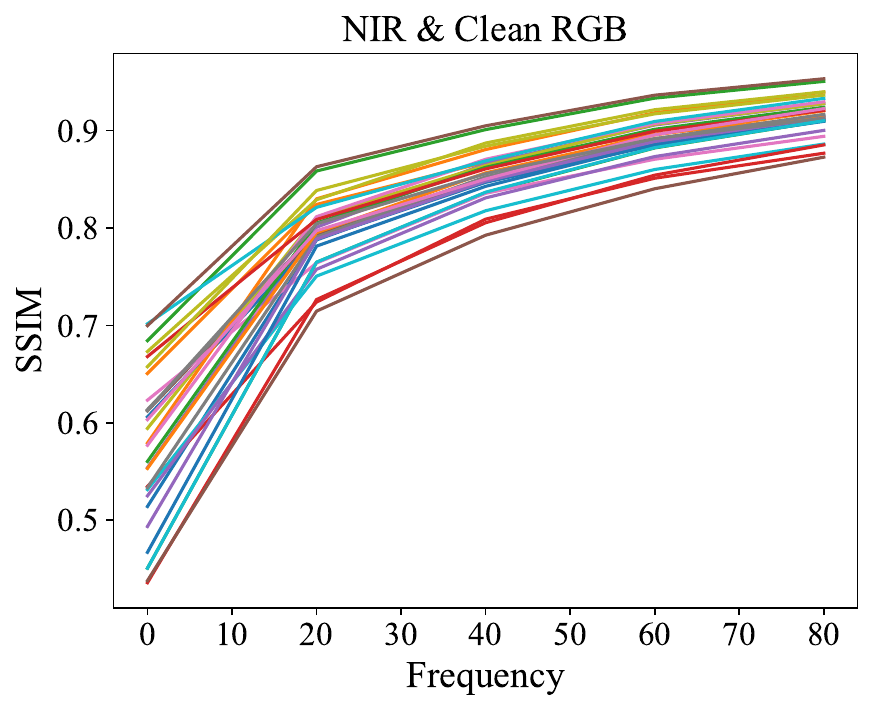} 
    \vspace{-6mm}
    \caption{}
    \label{fig:2a}
  \end{subfigure}
  \hspace{-2mm} 
  \begin{subfigure}[b]{0.235\textwidth}
    \centering
    \includegraphics[width=\linewidth]{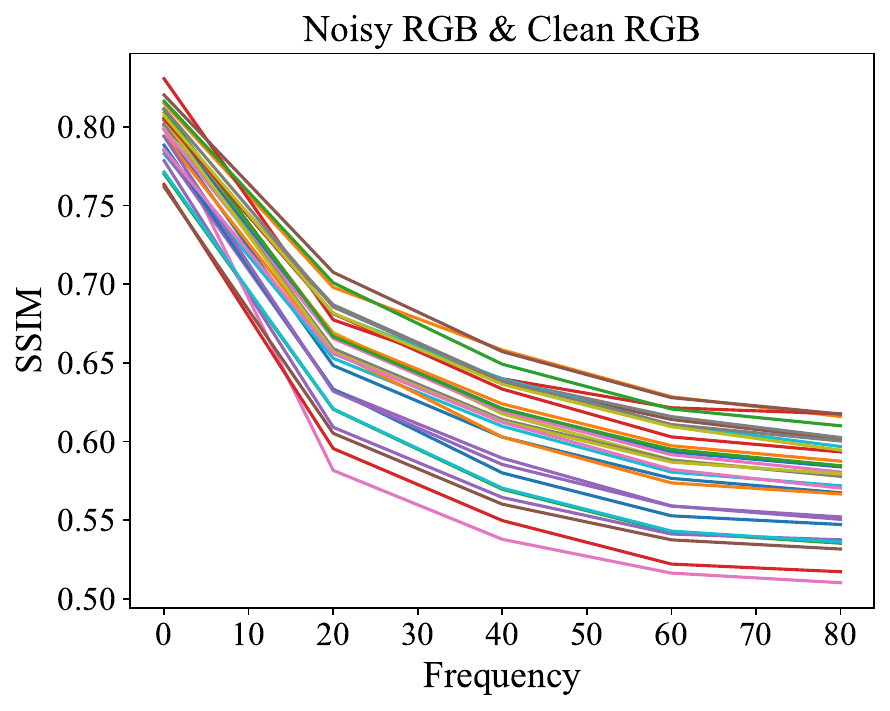}
    \vspace{-6mm}
    \caption{}
    \label{fig:2b}
  \end{subfigure}

  \begin{subfigure}[b]{0.45\textwidth} 
    \centering
    \includegraphics[width=\linewidth]{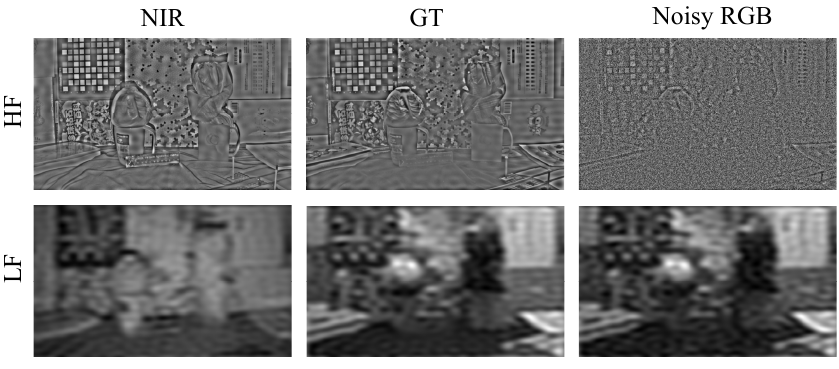}
    \vspace{-6mm}
    \caption{}
    \label{fig:2c}
  \end{subfigure}
  \vspace{-3mm} 
  \caption{The analysis of NIR and RGB frequency correlation. (a) The frequency correlation between noisy RGB and clean RGB, with the horizontal axis indicating the cutoff frequency of the high-pass filter. Different colored curves represent different scenarios. (b) The frequency correlation between NIR and clean RGB. (c) The visualization results of images output by a fixed frequency filter, with the first row showing high frequency (HF) part and the second row showing low frequency (LF) part.}
  \vspace{-3mm}
\end{figure}we first introduce the cross-field frequency correlation prior of NIR and RGB images. Based on the cross-field frequency correlation prior, a detailed frequency domain scheme is formulated, including the Frequency Dynamic Selection Mechanism (FDSM) and the Frequency Exhaustive Fusion Mechanism (FEFM) for thorough frequency exploitation.

\subsection{Cross-Field Frequency Correlation Prior}
To address the fusion of NIR and RGB images under spatial inconsistency, it is imperative to distinguish which information in NIR and RGB images is useful and which is redundant or even erroneous for denoising. We randomly select 30 pairs of NIR images, noisy RGB images, and clean RGB images from the existing RGB-NIR dataset and perform fourier transform on them. To facilitate the study of the correlation between NIR and RGB in the frequency domain, we apply a high-pass filter to the aforementioned three kinds of frequency domain images and convert the filtered results back to the spatial domain. We separately computed the Structural Similarity (SSIM)~\cite{wang2004image} between NIR/noisy RGB images and clean RGB images at different cutoff frequencies as a measure of frequency correlation $S$, as shown in Fig.~\ref{fig:2a} and Fig.~\ref{fig:2b}. The process can be described as
\begin{equation}
S=\mathrm{SSIM}(\mathcal{F}^{-1}(\mathrm{H}(F_{target})), \mathcal{F}^{-1}(\mathrm{H}(F_{gt})))
\end{equation}
where $\mathcal{F}^{-1}(\cdot)$ denote 2D-IDFT operations, $F_{target}$ and $F_{gt}$ represent NIR or noisy RGB and clean RGB after 2D-DFT transformation. $\mathrm{H}(\cdot)$ represents the filtering of the input image with a high-pass filter. The analysis results indicate that, in the main energy frequency, the image similarity between the input noisy RGB and clean RGB images decreases progressively as the frequency increases, while the image similarity between the input NIR image and clean RGB image increases from low to high frequency. Fig.~\ref{fig:2c} also shows that for noisy RGB images, their high frequency information is mainly composed of noise and lacks detail information, while the low frequency information more faithfully reflects the color in the original image, and NIR images are just the opposite. These cross-field Frequency correlation priors indicate that it is simple and efficient to distinguish useful information from NIR and RGB images in the frequency domain, which inspires us to propose a comprehensive frequency domain learning framework.
\begin{figure}[t!]
    \begin{center}
        \includegraphics[width=\linewidth]{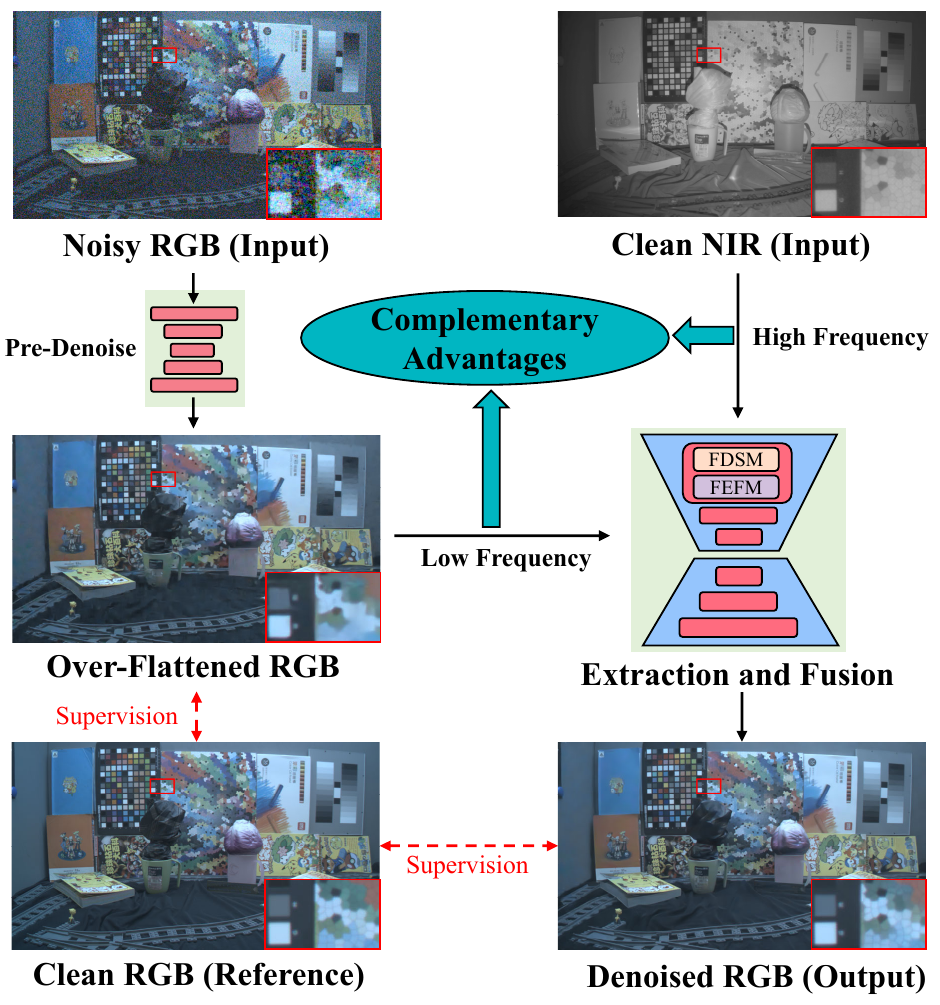}
    \end{center}
    \vspace{-6mm}
    \caption{The overall architecture of the proposed cross-field Frequency Correlation Exploiting Network (FCENet). In the first stage, pre-denoising is performed. In the second stage, NIR image is introduced for NIR-assisted image denoising.}
    \label{fig:3}
\vspace{-3mm}
\end{figure} 
\subsection{Overview Framework}
The proposed frequency domain fusion framework, as depicted in Fig.~\ref{fig:3}, includes two stages for the gradual restoration of images. Both stages incorporate an encoder-decoder architecture based on the U-Net~\cite{ronneberger2015u}, featuring two downsampling layers and two upsampling layers. Skip connections are employed between the encoder and decoder features. A Supervised Attention Module (SAM)~\cite{zamir2021multi} is utilized to connect features from the first stage to the second stage for further processing.
\begin{figure*}[t!]
    \begin{center}
        \includegraphics[width=\linewidth]{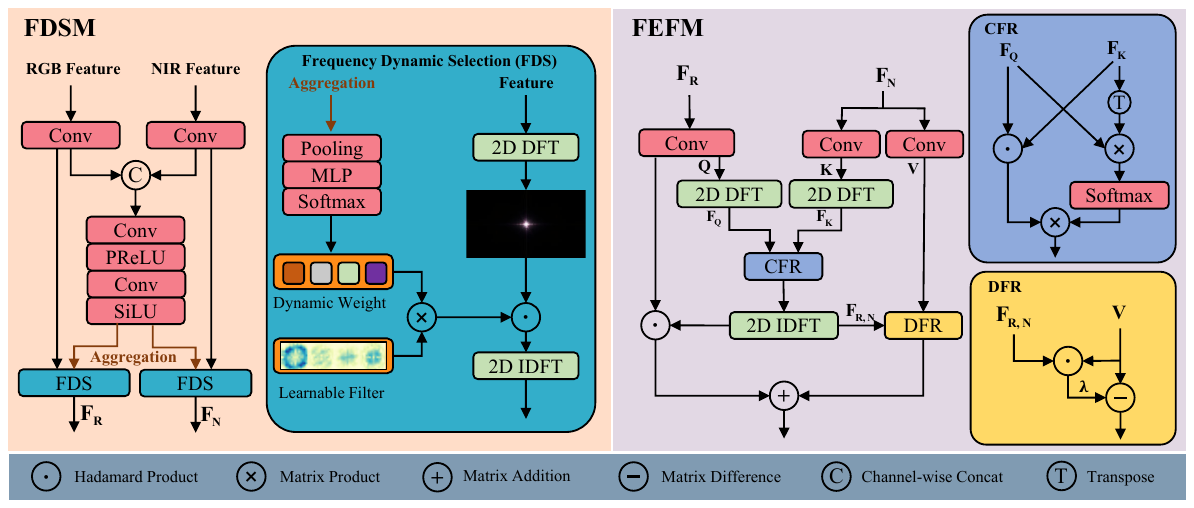}
    \end{center}
    \vspace{-5mm}
    \caption{The architecture of the proposed Frequency Dynamic Selection Mechanism (FDSM) and Frequency Exhaustive Fusion Mechanism (FEFM) for thorough frequency exploitation.}
    \label{fig:4}
\vspace{-4mm}
\end{figure*}

In the first stage, the input noisy RGB image undergoes preliminary denoising to reduce the modality gap with the NIR image. In the second stage, we perform multi-scale fusion of the encoded features of the NIR and RGB images. Specifically, at each scale of the encoder, NIR encoded features and RGB pre-denoise features are fed into the frequency domain fusion modules based on FDSM and FEFM, as shown in Fig.~\ref{fig:4}. 
\subsection{Frequency Dynamic Selection Mechanism}
The frequency correlation prior indicates that extracting cross-field complementary information from the frequency domain is efficient. For feature extraction, traditional convolutional kernels share weights across the entire feature map, lacking flexibility to address the numerous inconsistencies between NIR and RGB images, which affects the efficiency of extraction. We introduce frequency domain dynamic convolution to adapt to specific inputs and effectively extract complementary features.

Specifically, given the NIR features $\mathbf{N}\in\mathbb{R}^{C\times H\times W}$ and pre-denoised features $\mathbf{R}\in\mathbb{R}^{C\times H\times W}$, we concatenate them after layernorm and undergo convolution and the activate function to obtain a coarsely fused feature map, After that, we separate the fused feature map to obtain the aggregated features $\mathbf{A_{R}}$ and $\mathbf{A_{N}}$ for NIR and RGB individually. These aggregated features are used to generate dynamic filtering kernels to selectively extract useful features from the NIR and RGB input features. 

For dynamic filters, some works~\cite{zhou2021decoupled,ma2022generative} that generate deep separable and spatially variant dynamic filtering kernels are usually computationally intensive and time-consuming, and require a large amount of data. We employ a method similar to~\cite{chen2020dynamic,tatsunami2024fft} by generating dynamic weights to dynamically aggregate multiple parallel convolutional kernels related to the input, and performing this process in the frequency domain. Specifically, the aggregated features $\mathbf{A_{R}}$ and $\mathbf{A_{N}}$ are used to generate the weights for $k$ learnable frequency domain filters. For each feature channel, these filters are linearly combined to form a frequency domain global filter, which is used to select useful frequency domain information from the input NIR and RGB features. Finally, $\mathbf{F_{R}}$, $\mathbf{F_{N}}$ denoting the filtered useful features of NIR and RGB can be represented as
\begin{equation}
\mathbf{DF_{I}}=\mathrm{Softmax}(\mathrm{MLP}(\mathrm{Pooling}(\mathbf{A_{I}})))\cdot \mathbf{G}
\end{equation}\begin{equation}
\mathbf{F_{I}}=\mathcal{F}^{-1}(\mathcal{F}(\mathbf{I})\odot \mathbf{DF_{I}})
\end{equation}
where $\mathbf{DF_{I}}$ is frequency domain dynamic filter, and $\mathbf{I}$ represents $\mathbf{R}$ or $\mathbf{N}$. $\mathbf{G}\in \mathbb{R}^{M\times\ k}$ represents the combination of $k$ learnable frequency domain filtering kernels, and M is the spatial dimension of the convolutional kernel. $\mathcal{F}(\cdot)$ and $\mathcal{F}^{-1}(\cdot)$ denote the 2D-DFT and 2D-IDFT operations, respectively.
\subsection{Frequency Exhaustive Fusion Mechanism}
After the NIR and RGB feature maps pass through the FDSM, the useful features are retained. Next, to fuse the two inconsistent fields, we further integrate their information in the frequency domain and propose the FEFM, with this process driven by prior frequency correlations. Specifically, the proposed FEFM includes two key components: Common Feature Reinforcement (CFR) mechanism and Differential Feature Reinforcement (DFR) mechanism. CFR comprehensively explores the correlation between NIR and RGB features in the frequency domain to strengthen the modeling of common features, while DFR employs differential modeling to enhance key differential high frequency features.

\textbf{Common Feature Reinforcement Mechanism.} The convolution theorem establishes that the correlation or convolution of two signals in the spatial domain is equal to their element-wise product in the frequency domain. Leveraging this property, previous work~\cite{kong2023efficient} integrate the frequency domain into the self-attention mechanism, simplifying matrix multiplication into lightweight element-wise dot product operations. Although the dot product is efficient, it only calculates the point-wise correlation coefficients in the frequency domain, missing the long-range correlations in the frequency domain. To more comprehensively model the common features of NIR and RGB, we apply a point-wise convolution and a depthwise convolution to the input NIR and RGB feature maps $\mathbf{F_{R}}$, $\mathbf{F_{N}}$, to obtain encoded features $\mathbf{Q}{=}W_{d}^{Q}W_{p}^{Q}\mathbf{F_{R}}$, $\mathbf{K}{=}W_{d}^{K}W_{p}^{K}\mathbf{F_{N}}$, $\mathbf{V}{=}W_{d}^{V}W_{p}^{V}\mathbf{F_{N}}$. To facilitate the calculation of frequency domain correlations, we apply 2D-DFT to the input features to produce frequency cubes $ \mathbf{F_{Q}} $ and $ \mathbf{F_{K}}$. We simultaneously model the long-range correlation and point-wise correlation of NIR and RGB features in the frequency domain, resulting in a correlation map that integrates local similarity and long-range channel correlation within the frequency domain. The output of CFR can be represented as
\begin{equation}
\mathbf{F_{CFR}}=(\mathbf{F_{Q}}\odot \mathbf{F_{K}})\cdot \mathrm{Softmax}(\mathbf{F_{Q}}\cdot \mathbf{F_{K}^{\top}}/\alpha)
\end{equation}
where $\alpha $ is a learnable scaling \begin{figure*}[h]
\vspace{-3mm}
    \footnotesize
    \setlength\tabcolsep{1.8pt}
    \renewcommand\arraystretch{0.8}
    \begin{center}
        \begin{tabular}{c c c c c c c c}
             Noisy RGB  & NIR& Restormer & MNNet  & DVN & NIR-Restormer & \textbf{Ours} & Reference\\
            \includegraphics[width=.118\linewidth]{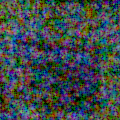} &
            \includegraphics[width=.118\linewidth]{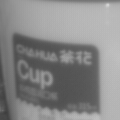} &
            \includegraphics[width=.118\linewidth]{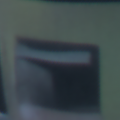} &
            \includegraphics[width=.118\linewidth]{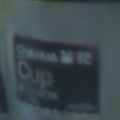} &
            \includegraphics[width=.118\linewidth]{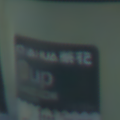} &
            \includegraphics[width=.118\linewidth]{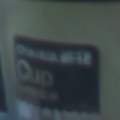}& 
            \includegraphics[width=.118\linewidth]{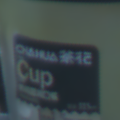} &
            \includegraphics[width=.118\linewidth]{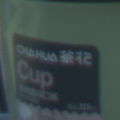}\\
             $\sigma = 2$ &PSNR& 31.44 & 31.50 & 33.65 & 32.63& \textbf{34.05}& GT\\
             \includegraphics[width=.118\linewidth]{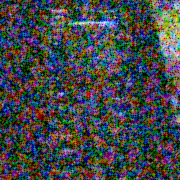} &
            \includegraphics[width=.118\linewidth]{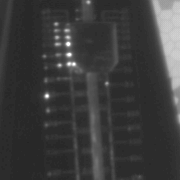} &
            \includegraphics[width=.118\linewidth]{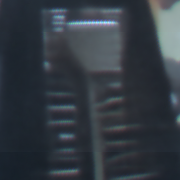} &
            \includegraphics[width=.118\linewidth]{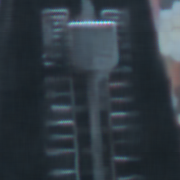} &
            \includegraphics[width=.118\linewidth]{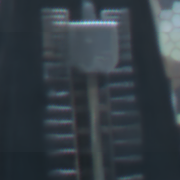} &
            \includegraphics[width=.118\linewidth]{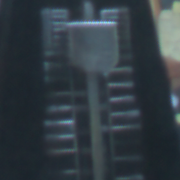}& 
            \includegraphics[width=.118\linewidth]{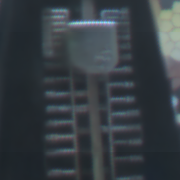} &
            \includegraphics[width=.118\linewidth]{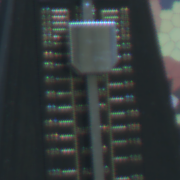}\\
             $\sigma = 4$ &PSNR&  27.02 & 27.41 & 27.28 & 27.33& \textbf{28.18}& GT\\
            \includegraphics[width=.118\linewidth]{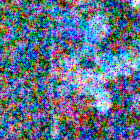} &
            \includegraphics[width=.118\linewidth]{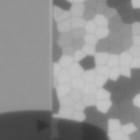} &
            \includegraphics[width=.118\linewidth]{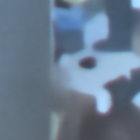} &
            \includegraphics[width=.118\linewidth]{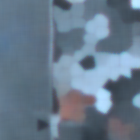} &
            \includegraphics[width=.118\linewidth]{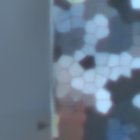} &
            \includegraphics[width=.118\linewidth]{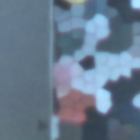}& 
            \includegraphics[width=.118\linewidth]{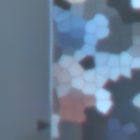} &
            \includegraphics[width=.118\linewidth]{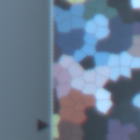}\\
              $\sigma = 6$ &PSNR& 26.37 & 27.02 & 27.55 & 28.54& \textbf{29.93}& GT\\             
        \end{tabular}
    \end{center}
    \vspace{-6mm}
    \caption{ The qualitative comparison among our FCENet and the state-of-the-art methods on the noisy RGB-NIR pairs from DVD\cite{jin2022darkvisionnet} with different noise levels. Our method achieves better results in detail recovery, artifact removal, and color restoration.}
    \label{fig:v_dvn}
\vspace{-1mm}
\end{figure*}
\begin{table*}[t]
\renewcommand{\arraystretch}{1.25}
\small
\tabcolsep=3.6mm
% \vspace{-1mm}
\caption{Comparison of different methods on the DVD test set with noise levels $\sigma = 2, 4, 6$.The best and second-best results are highlighted in boldface and underlined, respectively.}\label{table:1}
\vspace{-2mm}
\begin{tabular}{l|cc|cc|cc|cc}
\hline

\multirow{2}{*}{Methods} & \multicolumn{2}{c|}{$\sigma = 2$} & \multicolumn{2}{c|}{$\sigma = 4$} & \multicolumn{2}{c|}{$\sigma = 6$} &  \multicolumn{2}{c}{Complexity} \\ 
\cline{2-9} 
                         & PSNR$\uparrow$ & SSIM$\uparrow$ & PSNR$\uparrow$ & SSIM$\uparrow$ & PSNR$\uparrow$ & SSIM$\uparrow$  & FLOPs(G) & Params(M)\\ \hline
Uformer~\cite{wang2022uformer}       & 31.34& 0.949 & 29.12 & 0.927 & 27.58 & 0.908 & 9.60 & 20.63 \\ 
Restormer~\cite{zamir2022restormer}      & 31.30 & 0.949 & 29.20 & 0.928 & 27.76 & 0.910 & 36.06 & 26.13 \\ 
MPRNet~\cite{zamir2021multi}          & 31.78 & 0.951 & 29.37 & 0.929 & 27.84 & 0.910 & 133.51 & 15.74 \\ 
Xformer~\cite{zhang2023xformer}         & 31.41& 0.950 & 29.30 & 0.928 & 27.43 & 0.908 & 39.34 & 25.22 \\ \hline
CUNet~\cite{deng2020deep}        & 28.76 & 0.923 & 26.81 & 0.898 & 25.64 & 0.875 & \underline{6.72} & \textbf{0.44} \\ 
SANet~\cite{sheng2023structure}          & 30.13 & 0.938 & 27.84 & 0.917 & 26.41 & 0.901 & 25.01 & 5.38 \\ 
MNNet~\cite{xu2022model}            & 30.14 & 0.944 & 28.26 & 0.923 & 26.75 & 0.906 & 10.97 & \underline{0.76} \\ 
DVN~\cite{jin2022darkvisionnet}           & 31.50 & 0.955 & 29.62 & 0.940 & 28.26 & 0.927 & 48.30 & 6.97 \\ 
NIR-Restormer~\cite{xu2024nir}      & 31.81 & 0.957 & 29.81 & 0.942 & 28.31 & 0.928 & 45.31 & 30.26 \\ 
\rowcolor{mygray} 
FCENet-L (Ours)     & \underline{31.87} & \underline{0.960} & \underline{29.91} & \underline{0.946} & \underline{28.36} & \underline{0.934} & \textbf{5.71} & 1.16 \\
\rowcolor{mygray} 
\textbf{FCENet (Ours)}            & \textbf{32.43} & \textbf{0.963} & \textbf{30.37} & \textbf{0.950} & \textbf{28.78} & \textbf{0.939} & 19.52 & 4.44 \\ \hline
\end{tabular}
\vspace{-4mm}
\end{table*}parameter.

\textbf{Differential Feature Reinforcement Mechanism.} Traditional cross-attention~\cite{lin2022cat,wang2024in2set,guan2023mutual} structures are typically used to capture common features but do not effectively utilize differential information. Although this approach is correct in general image fusion frameworks, in this paper, the features of the two fields are considered useful after passing through the frequency domain selection mechanism. By observing the feature maps (Fig.~\ref{fig:7}), we find that the high frequency texture information in the NIR features is clear compared to the RGB feature maps. The common features obtained by weighted fusion through cross-attention remain somewhat blurry, which greatly affects the denoising effect. These differential high frequency features on NIR are crucial for image restoration. We adopt a differential cross-attention to complement the missing differential information, and the output of CFR can be represented as
\begin{equation}
\begin{split}
\mathbf{F_{DFR}}=(\mathbf{V}-\lambda\ \mathbf{V} \odot \mathcal{F}^{-1}(\mathbf{F_{CFR}}))
\end{split}
\end{equation}
where $ \lambda $ is a learnable scalar that dynamically controls the weights of two feature maps. Although DFR effectively models the differential high frequency information, the fused features lack some common features and background information. To further refine the fused features, we perform a dot product between the encoded features Q and the output of CFR, and add the result to the output of DFR as the final fused result.
\subsection{Loss Function}
We adopt the Charbonnier loss~\cite{charbonnier1994two}, and adding the frequency domain loss~\cite{cho2021rethinking}. The overall loss function is represented as
\begin{equation}
\begin{split}
\mathcal{L}=\mathcal{L}_{charbonnier}(\mathbf{X}_{1},\mathbf{T})+\mathcal{L}_{charbonnier}(\mathbf{X}_{2},\mathbf{T}) \\ 
+\alpha\cdot\mathcal{L}_{charbonnier}(\mathcal{F}(\mathbf{X}_{2}),\mathcal{F}(\mathbf{T}))
\end{split}
\end{equation}
where $\mathbf{X}_{1}$ and $\mathbf{X}_{2}$ are the outputs of pre-denoising sub-network and whole network, respectively. $\mathbf{T}$ is ground truth and $ \alpha\ $ is set as 0.1~\cite{cho2021rethinking}.
\section{Experiments}
% \vspace{-0.5mm}
In this section, we evaluate the proposed method on public datasets and compare it with other NIR-assisted image denoising methods.
\begin{figure*}[t!]
    \begin{center}
        \includegraphics[width=\linewidth]{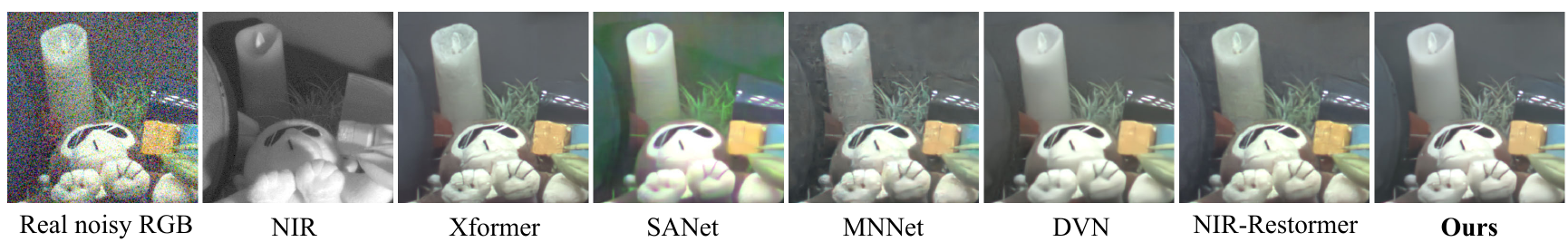}
    \end{center}
    \vspace{-6mm}
    \caption{Visual comparisons on real-world RGB/NIR image pairs. Our method demonstrates better visual results.}
    \label{fig:real}
\vspace{-3mm}
\end{figure*}
\subsection{Experimental Settings}
\textbf{Datasets.} We evaluate the NIR-assisted image denoising method on the DVD~\cite{jin2022darkvisionnet} and IVRG~\cite{brown2011multi} datasets. The DVD dataset consists of 307 high-resolution NIR-RGB image pairs, which are converted into 5000 image pairs of size $3 \times 256 \times 256$ for training and 1000 image pairs of size $3 \times 256 \times 256$ for testing. The IVRG dataset contains NIR and RGB image pairs from 9 different scenes. For each scene, as in~\cite{xu2022model}, experiments randomly select 30/8/15 pairs for training/validation/testing. \\
\textbf{Parameter settings.} The proposed network is implemented with Pytorch and trained using the Adam optimizer~\cite{kingma2014adam} ($\beta_1 = 0.9$, $\beta_2 = 0.999$) on patches of size $128 \times 128$ with a batch size of 16. For the DVD dataset, we simulate low-light conditions by randomly reducing the mean of the original images as in~\cite{jin2022darkvisionnet} and add Gaussian-Poisson mixed noise with a range of 1 to 16. The model is trained for 80 epochs. The initial learning rate is \(2 \times 10^{-4}\), and a cosine annealing strategy is used to gradually reduce the learning rate to $1 \times 10^{-6} $. For the IVRG dataset, following the settings in~\cite{xu2022model}, it is cropped into 16,374 training patches, and then additive Gaussian noise with random noise levels is used to simulate the target images, considering three noise levels in testing: $\sigma = 25$, 50, and 75. The model is trained for 100 epochs with an initial learning rate of $5 \times 10^{-4}$.
\begin{table}[t]
\renewcommand{\arraystretch}{1.25}
\footnotesize
\tabcolsep=0.78mm
\caption{Comparison of different methods on the IVRG test set with noise levels $\sigma = 25, 50, 75$. The best and second-best results are highlighted in boldface and underlined, respectively.}\label{table:2}
\hspace{-0.8mm}
\vspace{-3mm}
\begin{tabular}{l|cc|cc|cc}
\hline

\multirow{2}{*}{Methods} & \multicolumn{2}{c|}{$\sigma = 25$} & \multicolumn{2}{c|}{$\sigma = 50$} & \multicolumn{2}{c}{$\sigma = 75$}  \\ 
\cline{2-7} 
                         & PSNR$\uparrow$ & SSIM$\uparrow$ & PSNR$\uparrow$ & SSIM$\uparrow$ & PSNR$\uparrow$ & SSIM$\uparrow$ \\ \hline

Restormer~\cite{zamir2022restormer}        & 31.72 & 0.876 & 27.67 & 0.788 & 24.50 & 0.722 \\ 
Uformer~\cite{wang2022uformer}         & 31.97 & 0.878 & 27.83 & 0.793 & 24.58 & 0.726  \\ 
MPRNet~\cite{zamir2021multi}         & 31.89 & 0.878 & 27.80 & 0.794 & 24.57 & 0.729  \\ 
Xformer~\cite{zhang2023xformer}          & 31.98 & 0.879 & 27.86 & 0.794 & 24.61 & 0.731 \\ \hline
CUNet~\cite{deng2020deep}          & 31.37 & 0.876 & 27.78 & 0.812 & 24.64 & 0.764  \\ 
SANet~\cite{sheng2023structure}           & 31.54 & 0.885 & 28.13 & 0.825 & 24.81 & 0.785 \\ 
MNNet~\cite{xu2022model}           & 32.59 & 0.904 & 28.66 & 0.851 & 25.25 & 0.808  \\ 
DVN~\cite{jin2022darkvisionnet}             & 32.76 & 0.906 & 28.73 & 0.852 & 25.28 & 0.809  \\ 

NIR-Restormer~\cite{xu2024nir}      & \underline{32.86} & 0.906 & 28.77  & 0.853 & 25.30 & 0.811 \\ 
\rowcolor{mygray} 
FCENet-L (Ours)      & 32.85 & \underline{0.909} & \underline{28.80}  & \underline{0.857} & \underline{25.33} & \underline{0.816} \\
\rowcolor{mygray}
\textbf{FCENet (Ours)}            & \textbf{33.26} & \textbf{0.915} & \textbf{29.08} & \textbf{0.866} & \textbf{25.48} & \textbf{0.826}  \\ \hline
\end{tabular}
\vspace{-2mm}
\end{table}
\subsection{Experimental Results}
In our experimental evaluation of our method, we compare our results with nine models, including five state-of-the-art NIR-assisted RGB denoising methods: CUNet~\cite{deng2020deep}, SANet~\cite{sheng2023structure}, MNNet~\cite{xu2022model}, DVN~\cite{jin2022darkvisionnet}, and NIR-Restormer~\cite{xu2024nir}, and four state-of-the-art single-image denoising methods: MPRNet~\cite{zamir2021multi}, Restormer~\cite{zamir2022restormer}, Uformer~\cite{wang2022uformer}, and Xformer~\cite{zhang2023xformer}.  To further demonstrate the superiority of our method, we additionally provide the denoising results of a light version (-L) by
changing the initial feature map channel count from 64 to 36, which has only 1.16 M parameters. All compared methods are trained on the same training set as ours.\\
\textbf{Evaluations on the DVN dataset.} We first evaluate our method on the DVD dataset, with quantitative results shown in Table~\ref{table:1}. It can be observed that the proposed FCENet, and even FCENet-L, outperform all other single-image denoising methods and NIR-assisted denoising methods in terms of PSNR and SSIM with moderate computational and memory costs. The qualitative comparison in Fig.~\ref{fig:v_dvn} clearly illustrates that our method has advantages over other state-of-the-art methods in noise removal, color preservation, and detail recovery. For single-image denoising methods like Restormer, although they have achieved advanced denoising effects in low-noise scenarios, they still lack effective means to address detail texture loss in high-noise situations. For guided denoising methods, MNNet is unable to effectively handle the inconsistencies between NIR and RGB images. DVN strengthens consistent high frequency textures but loses critical differential high frequency information in NIR and sacrifices the authenticity of low-frequency color. NIR-Restormer 
\begin{figure}[t]
    \begin{center}
        \includegraphics[width=\linewidth]{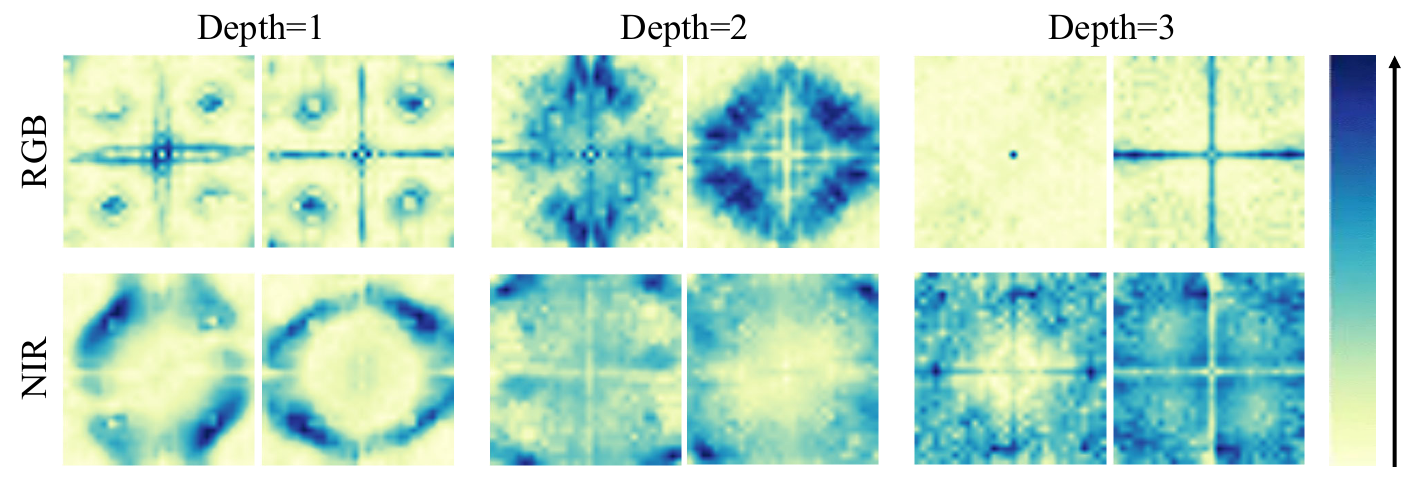}
    \end{center}
    \vspace{-4mm}
    \caption{Frequency domain dynamic filters on NIR and RGB side at different depths. The first row represents the frequency domain filters of RGB side, and the second row represents the filters of NIR side.}
    \label{fig:6}
\vspace{-5.5mm}
\end{figure} has good denoising effects but lacks effective utilization of inconsistent information, leading to over-smoothing and color distortion. Our method recovers detailed texture information while more faithfully preserving low-frequency color information in RGB images. This proves that the proposed method can efficiently utilize the complementary information of NIR and RGB images in the fusion through a series of frequency domain learning.\\
\textbf{Evaluations on the IVRG dataset.} We evaluate our method on the IVRG dataset. Following the previous method, we consider three noise levels in the test, including $\sigma = 25, 50$, and $75$. Quantitative results are shown in Table~\ref{table:2}. It can be observed that compared to the state-of-the-art single-image denoising algorithms, our method significantly improves the performance of image denoising, demonstrating the effectiveness of NIR images. Compared to existing NIR-assisted denoising methods, our method also has significant advantages.\\
\textbf{Evaluations on real-world experiment.} To further evaluate the performance of our model in real noisy environments, we conduct qualitative experiments on NIR-RGB data pairs in real low-light conditions. As shown in Fig.~\ref{fig:real}, the introduction of NIR images brings more detailed textures but also a large number \begin{table}[t]
\renewcommand{\arraystretch}{1.25}
\small
\tabcolsep=2.5mm
\centering
\caption{ Ablation study of various components of our method.}\label{table:ab_all}
\vspace{-2mm}

\begin{tabular}{c|ccc|cc}

\hline
 &Baseline & FDSM & FEFM &  PSNR$\uparrow$ & SSIM$\uparrow$ \\ 
\hline

(a) & \checkmark &  &  & 29.13 & 0.936   \\ 
(b) & \checkmark & \checkmark &  & 29.64 & 0.944  \\ 
(c) & \checkmark &  & \checkmark & 29.96  & 0.947   \\ 
\rowcolor{mygray} 
(d) & \checkmark & \checkmark & \checkmark & \textbf{30.37} & \textbf{0.950}   \\ 
\hline
\end{tabular}
\vspace{-0.5mm}
\end{table}
\begin{table}[t]
\renewcommand{\arraystretch}{1.25}
\small
\tabcolsep=3mm
\centering
\caption{Ablation on the branches of FDSM.}\label{table:ab_fdsm}
\vspace{-2mm}
\begin{tabular}{c|cc}
\hline
 Methods &  PSNR$\uparrow$ & SSIM$\uparrow$ \\ 
\hline
‌w/o FDSM & 29.96  &   0.947  \\ 
Spatial Dynamic Selection & 30.11 & 0.948  \\ 
\rowcolor{mygray} 
Frequency Dynamic Selection  &  \textbf{30.37} & \textbf{0.950}   \\ 
\hline
\end{tabular}
\vspace{-0.5mm}
\end{table}
\begin{table}[t]
\renewcommand{\arraystretch}{1.25}
\small
\tabcolsep=3mm
\centering
\caption{Ablation on the branches of FEFM.}\label{table:ab_fefm}
\vspace{-2mm}
\begin{tabular}{c|cc}
\hline
 Methods &  PSNR$\uparrow$ & SSIM$\uparrow$ \\ 
\hline
Sum & 29.64  &  0.944   \\ 
Cross-attention & 29.85 & 0.946  \\ 
DFR  &  30.15 & 0.948   \\ 
\rowcolor{mygray} 
DFR + CFR  &  \textbf{30.37} & \textbf{0.950}   \\
\hline
\end{tabular}
\vspace{-2mm}
\end{table}of inconsistencies. SANet fails to handle inconsistencies, resulting in a significant amount of artifacts, while MNNet, DVN and NIR-Restormer are unable to thoroughly denoise. In contrast, our method can eliminate inconsistencies while introducing more detailed information.
\subsection{Ablation Study}
We conduct ablation studies with FCENet on the DVD dataset under $\sigma = 4$. We compare two mechanisms proposed in this paper with a two-stage U-Net baseline, with the results shown in Table~\ref{table:ab_all}. In the U-Net baseline, cross-field images are fused through addition. By deploying FDSM and FEFM, the model achieves gains of 0.51dB and 0.83dB, respectively. Finally, combining all contributions significantly improves the denoising level compared to the baseline.\\
\textbf{Effect of FDSM.} Using FDSM allows for better extraction of complementary information from NIR and RGB while handling inconsistent regions. In Fig.~\ref{fig:6}, we show the frequency domain dynamic filter kernels in FDSM. It can be observed that the filter kernels processing NIR feature maps regularly enhance the learning of mid-to-high frequency information, while those processing RGB feature maps mainly learn low frequency information from the input images. This is consistent with the conclusions drawn from our frequency domain correlation priors. Additionally, we conduct experiments using spatial dynamic convolution in this module. The results in Table~\ref{table:ab_fdsm} ultimately prove the effectiveness of extracting relevant features from the frequency domain in NIR and RGB images and the stability of dynamic convolution in extracting inconsistent features.\begin{figure}[t]
    \begin{center}
        \includegraphics[width=\linewidth]{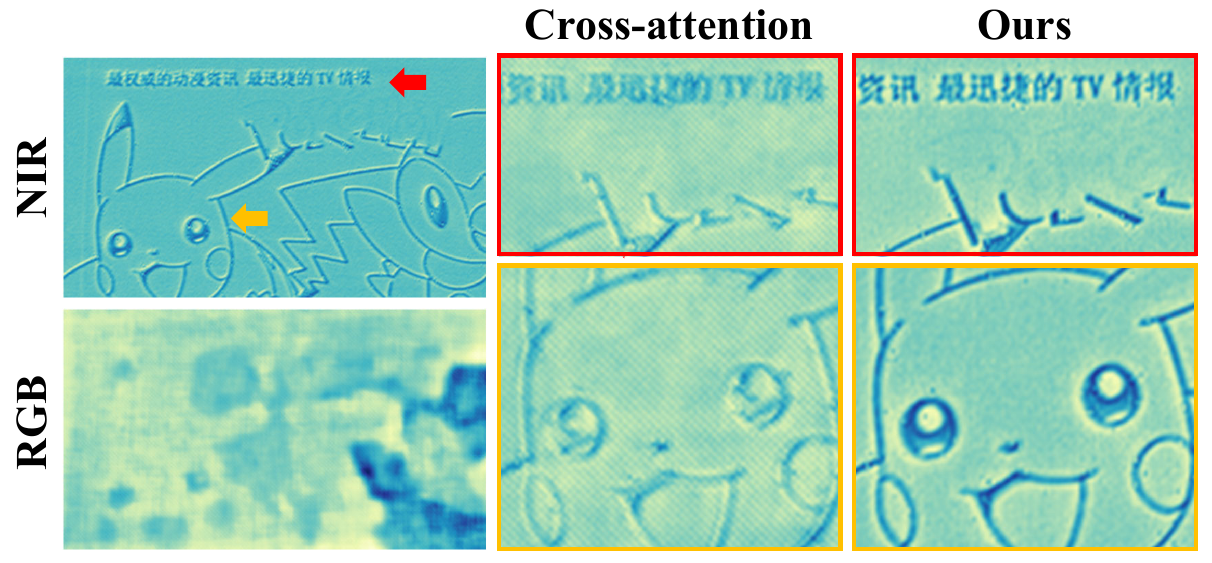}
    \end{center}
    \vspace{-4.5mm}
    \caption{ Feature maps obtained from two fields. The first column represents the input feature maps of NIR and RGB, the second column represents the output feature of cross-attention, and the third column is the output feature of FEFM.}
    \label{fig:7}
\vspace{-4mm}
\end{figure}\\
\textbf{Effect of FEFM.} Traditional cross-attention fusion modules only strengthen common features but neglect differential features. FEFM strengthens the learning of common features and enhances the learning of some high frequency differential information in NIR feature maps. Fig.~\ref{fig:7} shows the input NIR and RGB feature maps after frequency domain selection, the fused feature maps after cross-attention fusion, and the fused feature maps after FEFM. Compared to the NIR feature maps, the input RGB feature maps still lack high frequency features. It can be seen that the proposed FEFM strengthens common features while supplementing some differential texture information from NIR. We compare two mechanisms in FEFM, CFR and DFR, with traditional cross-attention fusion modules and direct feature summation. The results in Table~\ref{table:ab_fefm} prove effectiveness of our method in strengthening common features in frequency domain and modeling differential features.

\section{Conclusion}
In this paper, a cross-field frequency correlation exploiting network for NIR-assisted image denoising is proposed. Specifically, we first conduct a series of analyses on the characteristics of NIR and RGB images in frequency domain and obtain frequency correlation prior. Based on the prior, the frequency dynamic selection mechanism dynamically extracts complementary information from cross-field images in the frequency domain, and the frequency exhaustive fusion mechanism is proposed to enhance common features and differential high frequency features during fusion process. Extensive experiments demonstrate that the proposed method is significantly better than other algorithms. It is worth mentioning that existing methods may fail on significantly misaligned cross-field images. In future work, we plan to explore advanced methods such as feature point matching or incorporating deformable convolutional kernels to handle significantly misaligned cross-field images.
\section*{Acknowledgments}
This work is supported by the National Natural Science Foundation of China under Grants 62322204, 62131003, and the Key Laboratory of Target Cognition and Application Technology.

{
    \small
    \bibliographystyle{ieeenat_fullname}
    \bibliography{refs}
}


\end{document}